\documentclass[letterpaper, 10 pt, conference]{ieeeconf}
\IEEEoverridecommandlockouts
\usepackage{cite}
\usepackage{amsmath,amssymb,amsfonts}
\usepackage{subcaption}
\usepackage{graphicx}
\usepackage{float}
\usepackage{textcomp}
\usepackage{xcolor}
\usepackage{multirow}
\usepackage{array}
\usepackage{arydshln}
\usepackage{algpseudocode}
\usepackage[ruled,linesnumbered,vlined]{algorithm2e}
\usepackage{censor}
\usepackage{gensymb}
\usepackage[flushleft]{threeparttable}
\usepackage{booktabs,ragged2e}
\usepackage{blindtext}

\def\censorcolor{white}
\let\svcensorrule\censorrule
\renewcommand\censorrule[1]{%
\textcolor{\censorcolor}{\svcensorrule{#1}}}

\def\BibTeX{{\rm B\kern-.05em{\sc i\kern-.025em b}\kern-.08em
    T\kern-.1667em\lower.7ex\hbox{E}\kern-.125emX}}

\begin{document}

\title{\LARGE \bf Adaptive Hybrid Force–Vision Control with EMG-Derived References for Deformable Bottle Manipulation}

\author{ 
Jeffrey Lee, Geoffrey Hanks, Evangeline Bouganim, Yue Hu
\thanks{ 
Department of Mechanical and Mechatronics Engineering, University of Waterloo, Canada {\tt\footnotesize jeffrey.lee@uwaterloo.ca, gthanks@uwaterloo.ca, ebouganim@uwaterloo.ca, yue.hu@uwaterloo.ca}
}
\thanks{
We acknowledge the support of the Natural Sciences and Engineering Research Council of Canada (NSERC) through the CREATE ADVENTOR program under grant 584985 and NSERC PGS-D under grant 600034-2025.
}
}

\maketitle

\begin{abstract}
    Robotic automation of laboratory tasks has become a growing interest to remove researchers from hazardous environments and repetitive procedures. However, even equipment as simple as squeeze bottles poses a challenging manipulation problem due to its deformability. This paper presents an adaptive hybrid vision-force controller for robotic grippers to effectively manipulate laboratory squeeze bottles to dispense liquid. This approach integrates force and vision feedback according to the interaction state, enabling effective squeezing while accounting for bottle deformation and recovery. The proposed controller is validated by comparing its performance with a pure force controller on 12 combinations of conditions consisting of 2 target dispense volumes, 2 plastic squeeze bottles, and 1 to 3 squeezes used to dispense. Human demonstrations of these operating conditions were collected using electromyography (EMG) sensing mapped to a desired grip force, which was then used as the force reference for the controllers to squeeze the bottles. Results demonstrated that the proposed adaptive hybrid vision-force controller achieved more accurate liquid dispensing with a mean absolute error (MAE) of 6.54 mL compared to a MAE of 15.92 mL from the pure force controller, while also replicating the human action recorded from the EMG signal more closely.

\end{abstract}

\section{Introduction}

There is an increasing interest in using robotic manipulators for laboratory automation, as they can automate repetitive tasks and limit contact between researchers and hazardous materials through automation and remote operation \cite{THUROW2026100373}. Laboratory settings demand that robots perform experiments that require precise, accurate, and repeatable manipulation of deformable objects such as squeeze bottles, used for dispensing solutions and cleaning equipment. However, current robotic systems struggle to reliably perform contact-rich manipulation on deformable objects in dynamic environments that require high repeatability and accuracy \cite{Arriola-Rios2020} \cite{Cooper2025}. Squeeze-bottle manipulation is particularly challenging as it requires coordinated regulation of grip force and object deformation, especially when multiple squeezes are required, and the bottle must recover between successive interactions.

\begin{figure}
    \centering
    \includegraphics[width=1.0\linewidth]{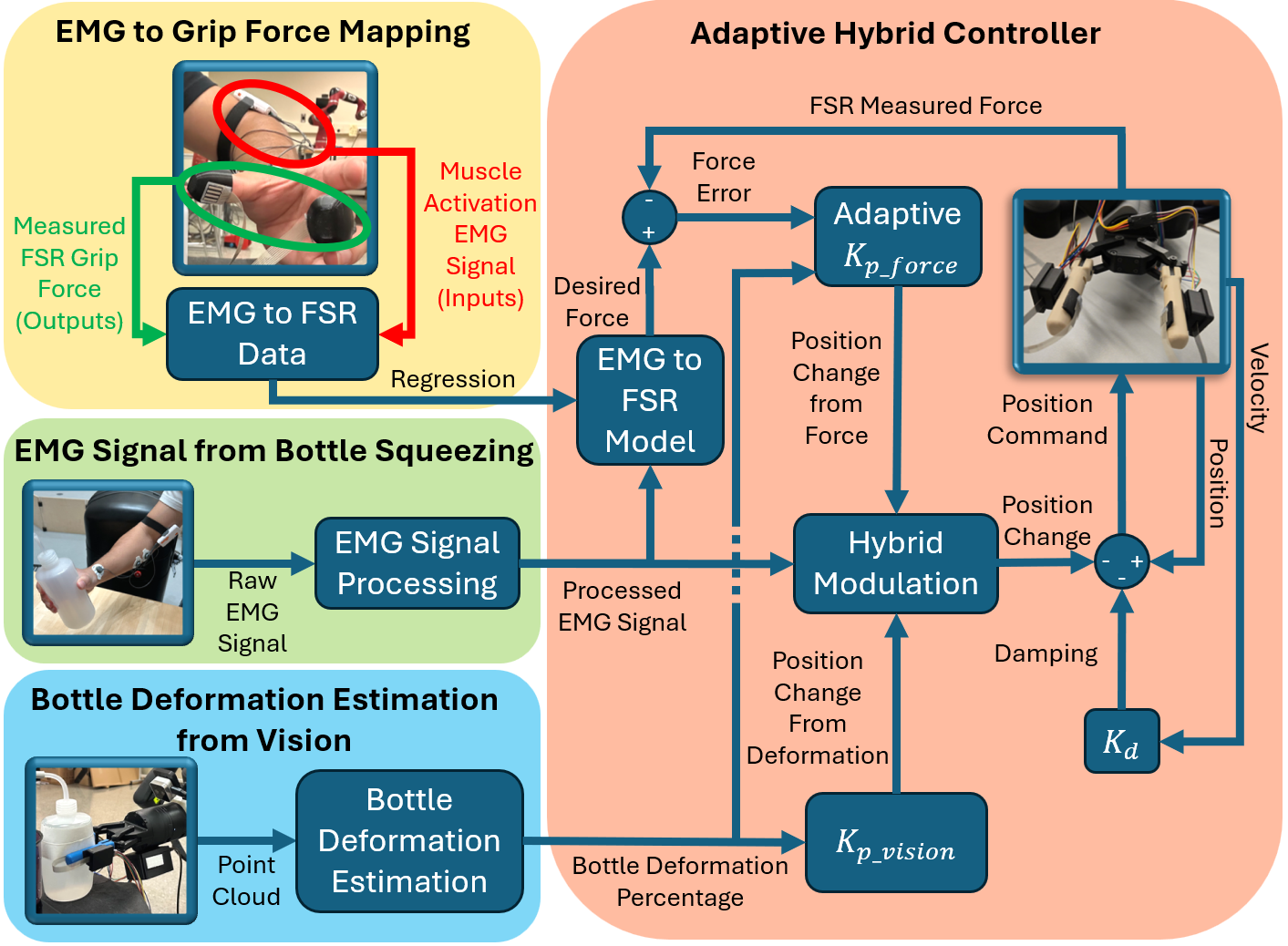}
    \caption{The adaptive hybrid force-vision controller framework detailing the EMG to grip force mapping, EMG signal processing, and bottle deformation estimation using vision inputs into the adaptive hybrid controller.}
    \label{fig:framework}
\end{figure}

Tactile sensors directly measure the grip force being applied by a robot manipulator, providing key feedback regarding the state of deformable objects \cite{cong2024tacfr}. However, experiments revealed that force feedback alone was insufficient to represent the state of a wash bottle due to similar force measurement readings at different levels of deformation. In particular, during the ``release phase", after water has been dispensed and the bottle is returning to its original form, commanding a low grip force was not sufficient to allow the bottle to return to its original shape. Therefore, machine vision can provide complementary information about the object state by directly estimating levels of deformation \cite{10.3389/frobt.2020.00073}, motivating the integration of force and vision for squeeze bottle manipulation.

In addition to sensing the state of the object, force-based manipulation requires a reference specifying the desired interaction force. Rather than manually defining this reference, human demonstrations can provide reference force profiles to replicate desired squeezing behaviours. This serves to both create desirable squeezing profiles, and allow for potential direct teleoperation of the system. Direct operator control allows for increased safety and adaptability, which is critical in laboratory settings to modify experiments and operate in human-centered lab environments \cite{Gentile2023}. Operator inputs can be captured using EMG sensors, which provide a non-invasive method of measuring grip force through muscle activation \cite{xue2023} and do not interfere with manipulation contact, allowing for more intuitive demonstrations to be collected. In this work, EMG signals recorded during human demonstrations are mapped to desired robot grip-force profiles, providing human-derived references for reproducing the demonstrated squeezing behaviour. 

The multiple sensing modalities allow both the intended interaction force and the changing deformation state of the object to be incorporated into the manipulation task. Building on this multimodal information, this paper presents two main contributions:
1) an EMG-driven adaptive hybrid force-vision controller for deformable squeeze-bottle manipulation in laboratory automation, combining reference force profiles from human demonstrations with grip force measured using force sensitive resistors (FSRs) and visually estimated object deformation, as shown in Fig. \ref{fig:framework}. The controller adaptively integrates force and deformation feedback to regulate squeezing while allowing the bottle to recover between successive interactions; and
2) a comprehensive set of experimental validations on the TIAGo Pro robot (PAL Robotics, S.L., Spain), demonstrating that the proposed controller can more accurately replicate human demonstrations and achieve improved dispensing accuracy when compared to a force-only controller, across different bottle sizes, target amounts, and number of squeezes. 



\section{Background}

\subsection{Electromyography Usage in Robotics}
Electromyography (EMG) measures the electrical signals generated from neuromuscular activity, and can be recorded non-invasively using surface electrodes (sEMG) \cite{gohel2020review}. Since processed EMG amplitude is correlated with force generated by the muscle, EMG signals can be used to estimate applied force during isometric tasks \cite{coggshall1970emgforce}. This enables the collection of human grasping demonstrations without interfering with the manipulated object or restricting natural hand motion.

EMGs are widely used to estimate human force through muscle activity for human-robot interaction. Xue \emph{et al.} and Gowtham \emph{et al.} used EMGs to measure muscle activity in the forearm and predicted grip force for position control of a robotic gripper \cite{xue2023, 9198439}. Geetha \emph{et al.} used EMGs for arm gesture recognition to control a mobile manipulator \cite{KS2025104563}. Cheon \emph{et al.} used EMGs to actuate a robotic glove to augment the user's grasping force \cite{9305235}. Dosen \emph{et al.} used EMGs for feedback control to augment the gripping force of a myoelectric prosthesis \cite{dosen2015emg}. While these works demonstrate the potential of EMGs for human-in-the-loop control, they primarily focus on rigid-object manipulation or force augmentation. EMG signals provide information about human force but do not directly capture changes in the manipulated object's state during interaction. Therefore, their application to contact-rich deformable object manipulation remains underexplored.

\subsection{Deformable Object Manipulation in Robotics}
Deformable objects present a unique challenge for robot manipulation tasks since the object's state changes as it is manipulated. As Zhu \emph{et al.} discuss in a review, deformable object manipulation remains an active area of research needing development across hardware design, sensing, deformation modeling, planning, and control \cite{Zhu_deform}. 
Some works in literature rely on model-based estimation for a soft object’s physical properties \cite{Wada_ICRA}, \cite{Wada2000} \cite{Arriola-Rios2020}. However, these models often require extensive calibration and are difficult to generate, limiting their scalability when manipulating objects with large variations in material properties. 

Some approaches avoid manually creating a model by using force feedback and learning. Lee \emph{et al.} learned force-based manipulation from human demonstrations of end-effector poses and the applied forces at each pose \cite{lee_icra}. Sanchez \emph{et al.} used measured contact forces while deforming objects to continuously update a deformation model  \cite{10.3389/frobt.2020.00073}. Zhong \emph{et al.} proposed a regressor-based approach in which using force feedback, the robot can continuously improve its model of a linear deformable object’s behaviour instead of using a fixed model \cite{zhong_robio}. However, force feedback alone provides limited information about the overall state of the object outside of the force sensor contact points.

Other approaches adopt model-free methods by using computer vision to track an object's deformations. Maitin-Shepard \emph{et al.} used vision-based grasp point detection to detect the corners of cloth to autonomously fold a towel \cite{maitin-shepard}. Sun \emph{et al.} used RGB-D images of clothing on an active stereo robot head to interpret surface wrinkles in clothing for dual-arm manipulation \cite{sun_icra}. Lee \emph{et al.} demonstrated in two studies that using RGB-D images and point clouds to track soft-object deformations for adaptation of impedance control parameters improved both the accuracy and precision of the manipulation of doughs and soft tissues \cite{VAISI, VAIRO}. 

While model-based methods have demonstrated effective deformable object manipulation, scaling such approaches to laboratory automation remains challenging. In laboratory environments, robots may interact with a wide variety of deformable tools and containers, making it impractical to manually construct or learn object-specific physical models for each object. Therefore, this work adopts a model-free vision-based approach to estimate deformation during manipulation, providing object-state information that complements force feedback.

\subsection{Hybrid Force-Vision Controllers}
A central theme in deformable object manipulation is that both force feedback and object deformation information is necessary for the task. Hybrid force-vision control leverages both sensing modalities to fuse contact-rich data with overall object state. 
Leite \emph{et al.} and Antonio \emph{et al.} used hybrid vision-force controllers to interact with and estimate unknown surface geometry using a fixed camera and wrist-mounted force sensors \cite{leite_icra}, \cite{antonio_sage}. Sheng \emph{et al.} extended this work to apply hybrid vision force control on deformable surfaces using variable impedance control to handle the non-linear and dynamic stiffness \cite{sheng_m2vip}.


These works on hybrid force-vision control primarily focus on autonomous manipulation and do not consider the integration of human manipulation strategies into deformable object interaction. Humans naturally combine muscle activation, contact force, and visual feedback to adapt their actions during manipulation. However, the integration of these sensing modalities for reproducing human manipulation behaviours in deformable object interaction remains underexplored. Therefore, this work investigates the integration of EMG-derived human force references with force and vision feedback for human-like deformable object manipulation.

\section{Methodology}
\label{sec:Methodology}

\subsection{Framework}

The framework proposed in this paper consists of four main components, as shown in Fig.\ref{fig:framework}. First, an EMG to grip force mapping using FSRs must be established to translate EMG signals to force signals to be used as the force reference for the controller. Second, EMG signal processing must also be performed to extract usable EMG envelopes. Third, a point cloud vision system is used to track the deformation of the squeeze bottles for adaptation in the hybrid controller. Lastly, the adaptive hybrid controller combines these signals to replicate the bottle squeezing action from the EMG-derived human demonstration. 

\subsection{EMG Placements and Signal Processing}
\label{sec:EMGPlacements}
\begin{figure}
    \centering
    \includegraphics[width=0.85\linewidth]{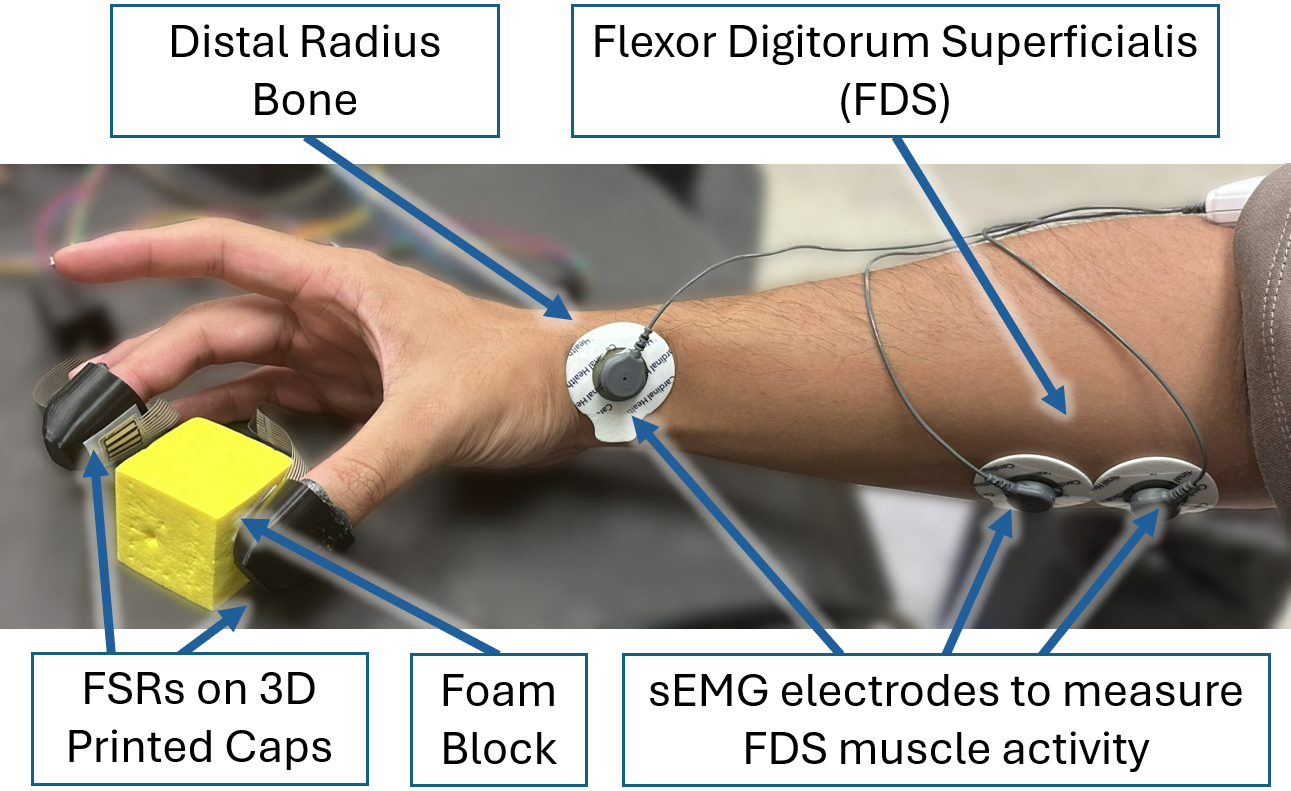}
    \caption{FSRs and sEMG electrode placements on human operator's arm for FSR to EMG mapping data collection.}
    \label{fig:human-data-collection-setup}
\end{figure}

The Shimmer3 EXG sensor units (Shimmer Research Ltd., Ireland) were used for this work. The EMG signal was collected at 1024 Hz with a gain of 12 on the sensor unit. The signal was rectified and filtered using a 4th-order Butterworth bandpass filter at 20-500 Hz to remove noise and movement artifacts to extract the EMG envelope\cite{ABCofEMG}. The placement of sEMG to measure muscle activity to predict grip force is an extensively researched topic. The flexor digitorum superficilais (FDS) is a muscle in the forearm that flexes the middle phalanges (long bones in fingers) of the medial 4 digits, and reinforces flexion against resistance (squeezing) \cite{okafor2022anatomy}. Literature has seen success in estimating grip force using sEMGs on the FDS \cite{11021574, xue2023, 9305235}. From literature and experimental trials of bottle squeezing, sEMG electrodes were placed as seen in Fig. \ref{fig:human-data-collection-setup}. For this work, a ground terminal was placed on the distal radius bone, and the positive and negative terminals were placed approximately on the centre of the FDS. 

\subsection{Gripper Design}
\begin{figure}
    \centering
    \includegraphics[width=0.85\linewidth]{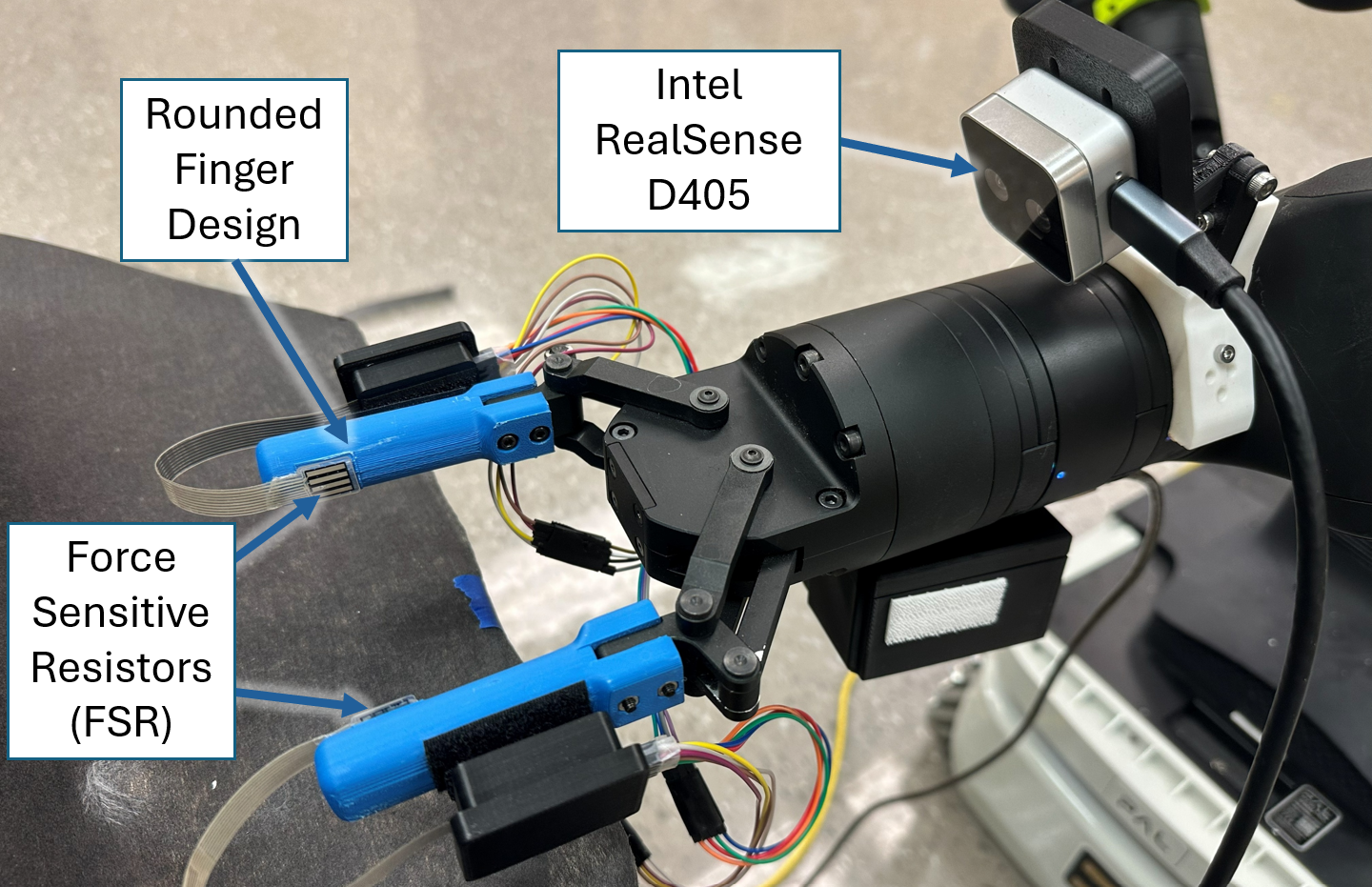}
    \caption{End-effector with the Intel RealSense D405 camera at the wrist and custom rounded-finger gripper equipped with force sensisitive resistors.}
    \label{fig:end-effector}
\end{figure}

The two-finger gripper used in this paper has standard flat-surface fingers. Therefore, the concave deformation of the bottle when squeezed causes sensors on flat grippers to lose contact with the bottle's surface. As such, custom gripper fingers had to be designed to ensure the force sensor's contact and accurate reading throughout the squeezing task. Leveraging the flexible FSRs, curved surface fingers were designed to ensure reliable contact between the sensor and bottle, even during deformation. This rounded finger design and placement of the FSRs on the gripper's fingers can be seen in Fig. \ref{fig:end-effector}. Furthermore, the FSR sensor has a 4x4 grid of tactile pressure sensing nodes. With a FSR sensor placed on each finger of the gripper, the average of the 32 node readings were used as the measured gripping force. 

\subsection{EMG to FSR Mapping}
To transfer human squeezing force via muscle activations, a map between EMG signals to grip force must be established. To collect the data for this mapping, EMG signals of the FDS muscle activity and FSR signals of the squeezing force were collected simultaneously during squeezing of a foam block. The sEMG electrodes were placed as described in sec. \ref{sec:EMGPlacements} and FSRs were fixed onto rigid 3D-printed caps on the user's thumb and middle finger (see Fig. \ref{fig:human-data-collection-setup}) to best replicate the gripping of the designed rigid two-fingered robotic gripper.

\subsection{Bottle Deformation Estimation Using Computer Vision}
The designed controller performs its adaptation based on a visual estimation of the squeeze bottle's deformation relative to its undeformed size. This allows deformation estimation to be independent of bottle size. A depth camera placed on the manipulator's final link, angled downward at $35\degree$, is used to capture the point cloud of the bottle during the squeezing task. This places the centre of the captured image at the centre of the gripper, as seen in Fig. \ref{fig:end-effector}.  

\begin{figure}
    \centering
    \includegraphics[width=0.7\linewidth]{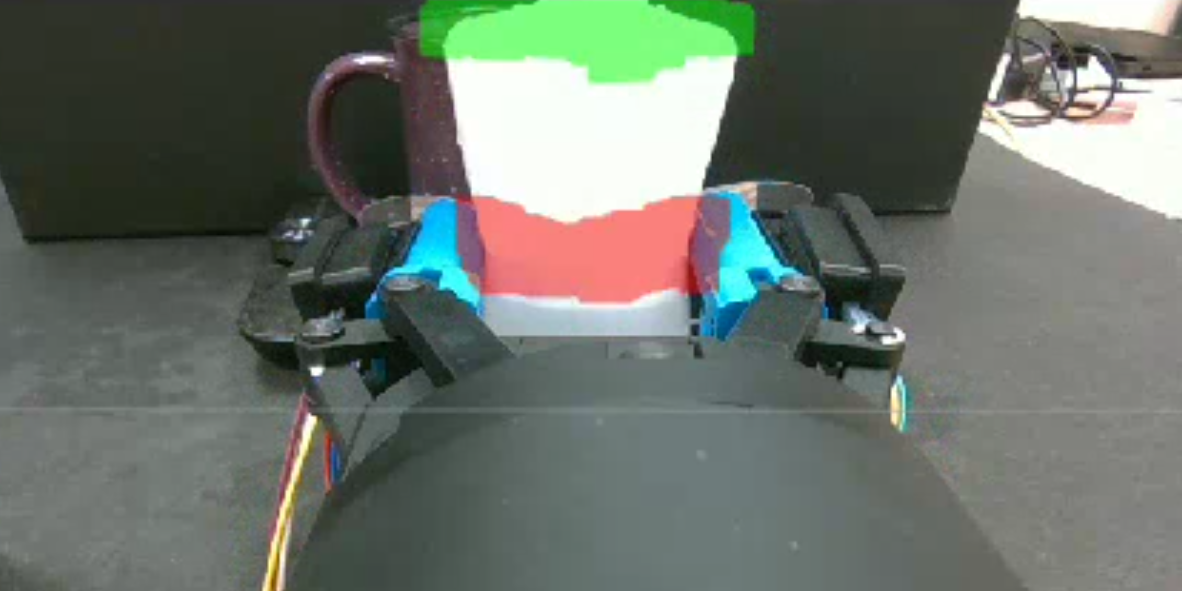}
    \caption{Vision deformation estimation system with point cloud measurement points overlaid. Green points represent those used for the undeformed bottle size reference and red points represents the deformed measurement.}
    \label{fig:vision_deformation}
\end{figure}

The point cloud processing to estimate the bottle deformation was done using the Point Cloud Library (PCL) with the following steps \cite{Rusu_ICRA2011_PCL}: 

\begin{enumerate}
    \item Extract bottle points from the captured point cloud using color-based clustering and the known gripper location in the camera frame.
    
    \item Transform the point cloud to align the bottle's cross-section with the $x$-$z$ plane.
    
    \item Estimate the undeformed bottle width by averaging the widest $x$-$z$ slice over a 1~cm height range (green region in Fig.~\ref{fig:vision_deformation}).
    
    \item Estimate the deformed bottle width by averaging the thinnest $x$-$z$ slice over a 1~cm height range (red region in Fig.~\ref{fig:vision_deformation}).
    
    \item Calculate deformation as the percentage reduction from the undeformed to deformed width, enabling size-independent measurement.
\end{enumerate}

\subsection{Adaptive Hybrid Control}
To replicate human bottle-squeezing behaviour using a position-controlled robotic gripper, EMG signals recorded during human demonstrations were mapped to desired FSR readings. This desired squeezing force profile was then used as the force reference for the position based force controller. However, force control alone was found to reproduce the squeezing phase more effectively than the release phase. In particular, the gripper could close in response to increasing EMG-derived force commands, but often failed to reopen when the EMG signal decreased, even when the mapped desired force indicated relaxation. This limitation motivated the introduction of a hybrid adaptive controller that combines EMG-derived force feedback with vision-based estimation of bottle deformation, as illustrated in Fig.~\ref{fig:framework}.

The controller is designed such that force feedback dominates during active squeezing, when the EMG signal is high, while bottle-deformation feedback becomes dominant when the EMG signal is low. This allows the gripper to respond not only to the desired squeezing force but also to the deformation state of the bottle, particularly during grip relaxation and reopening. The desired gripper position is given by:
\begin{equation} 
\begin{split} 
    x_d = x &- \sigma K_{p,\mathrm{f},\mathrm{adapt}} (f - f_d) \\ &- (1-\sigma) K_{p,\mathrm{deform}} B_{\mathrm{deform}} - K_d \dot{x}. 
\end{split} 
\label{eq:control_eq} 
\end{equation}

where $x_d$ is the desired gripper position, $x$ is the measured gripper position, and $\dot{x}$ is the measured gripper velocity. The desired force $f_d$ is obtained from the EMG-to-FSR mapping, while $f$ is the measured FSR force at the gripper. The term $B_{\mathrm{deform}}$ represents the bottle deformation estimated from vision. The modulation factor $\sigma$ varies with the EMG signal and determines the relative contribution of force and deformation feedback: larger values of $\sigma$ place greater emphasis on force control, whereas smaller values increase the contribution of deformation feedback. When using pure force control to evaluate the impact of the machine vision, $\sigma$ was set to 1 to remove the vision based compensation and only utilize force feedback. The gain $K_{p,\mathrm{f},\mathrm{adapt}}$ is the adaptive proportional gain applied to the force error, $K_{p,\mathrm{deform}}$ is the proportional gain associated with bottle deformation, and $K_d$ is the damping gain applied to the gripper velocity. 

Additional experiments showed that greater squeezing force was required as the deformation of the bottle increased. To account for this behaviour, the proportional gain applied to the force error was made adaptive on the estimated bottle deformation. The adaptive force gain is defined as:
\begin{equation} 
\begin{split} 
    K_{p,\mathrm{f},\mathrm{adapt}} = K_{p,\mathrm{f}} + K_{p,\mathrm{f},\mathrm{deform}}max(B_{\mathrm{deform}}-B_{\mathrm{nom}},0)    
\end{split} 
\label{eq:adaptive_force_gain} 
\end{equation}
where $K_{p,\mathrm{f}}$ is the nominal proportional gain on the force error, and $K_{p,\mathrm{f},\mathrm{deform}}$ determines how strongly the force gain increases with bottle deformation. The parameter $B_{\mathrm{nom}}$ defines the deformation threshold above which the additional gain becomes active. Consequently, the controller increases its force-feedback gain only when the estimated bottle deformation exceeds the nominal deformation level, enabling stronger squeezing action when larger forces are required.

\section{Experiments and Results}
\subsection{Experimental Setup}
\begin{figure}
    \centering
    \includegraphics[width=0.9\linewidth]{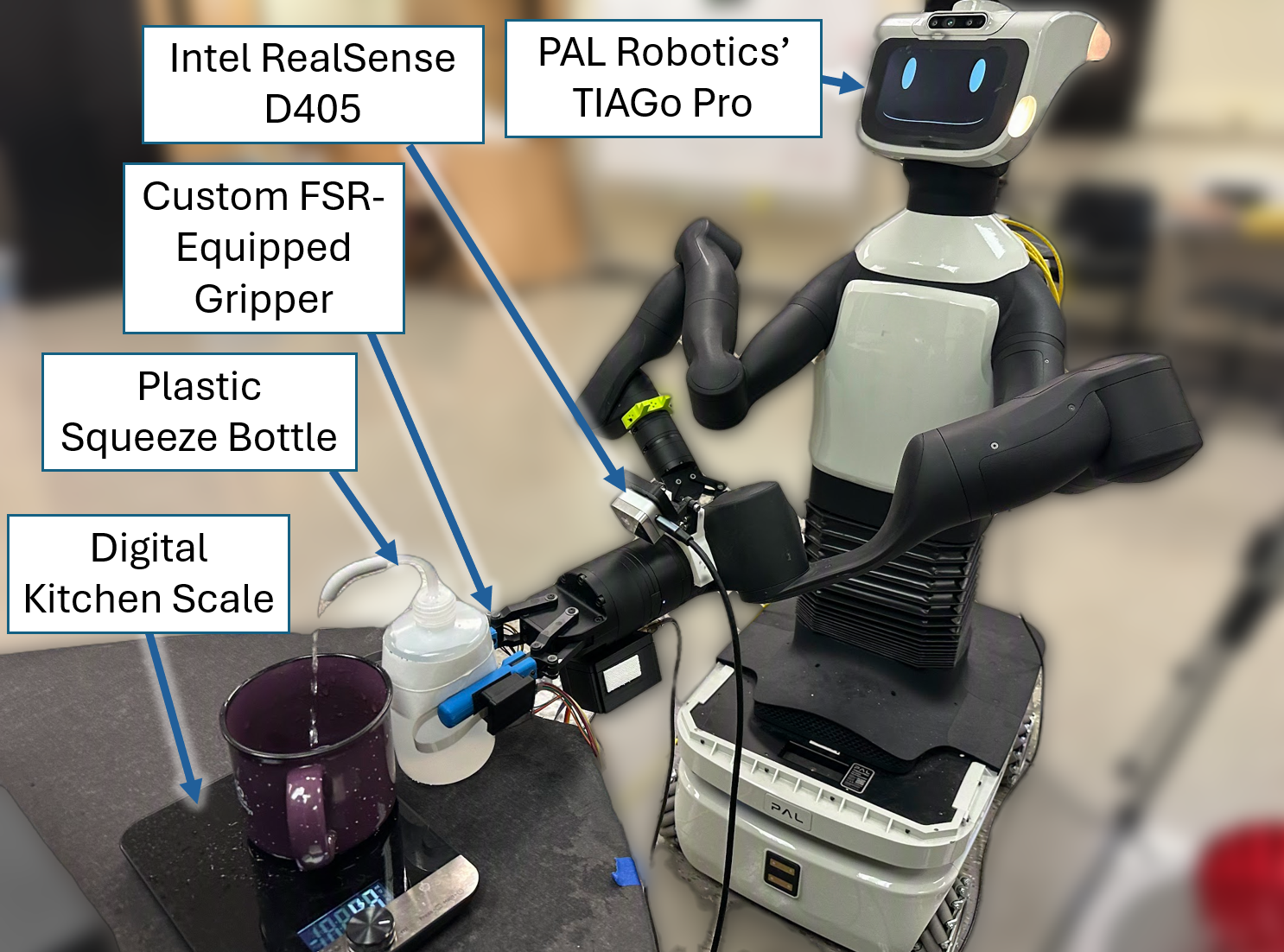}
    \caption{Experimental setup of the robotic bottle squeezing task using the TIAGo Pro, an Intel RealSense D405 camera, and an FSR-equipped gripper used to squeeze a plastic bottle to dispense water into a cup on a digital scale.}
    \label{fig:exp_setup}
\end{figure}

 The experimental setup for the adaptive hybrid controlled robotic bottle squeezing task is depicted in Fig. \ref{fig:exp_setup}. The TIAGo Pro's gripper (position-controlled from 0 (opened) to 1 (closed)) was equipped with the custom designed fingers using RX-M0404S FSR sensors (Roxi Electronic Technology Co., Ltd., China) and an Intel RealSense D405 depth camera (Intel Corporation, USA) was attached at the end-effector's final link for the vision-based bottle deformation estimation. Commercial plastic wash bottles were used as the squeeze bottles, and a digital scale with $\pm$1mL accuracy was used to measure the water dispensed from the experiments. 
 
\begin{figure}
    \centering
    \includegraphics[width=0.9\linewidth]{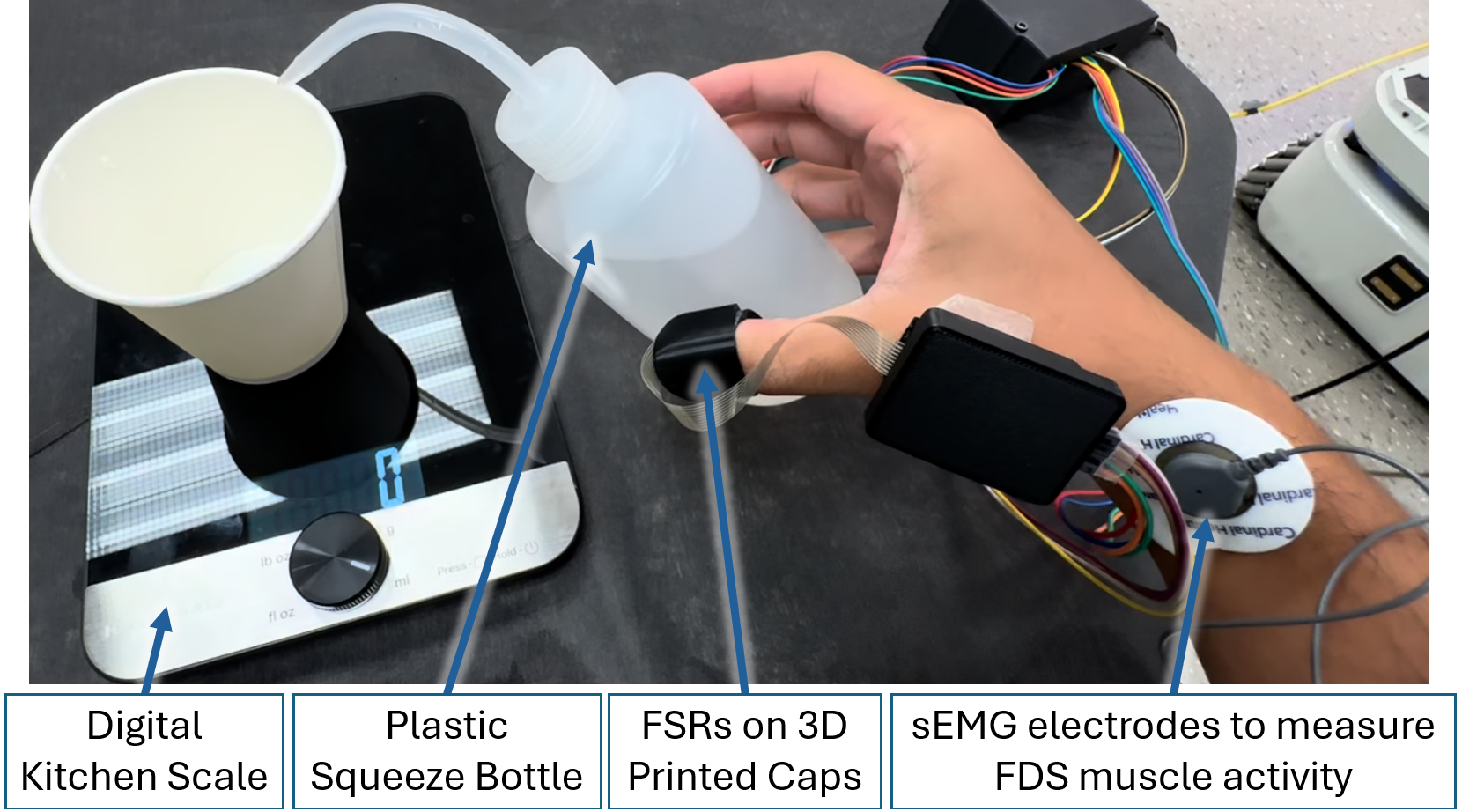}
    \caption{Experimental setup to collect human demonstrations of bottle squeezing. FSR finger-caps are placed on the thumb and middle finger and sEMG electrodes are placed on the FDS and distal radius bone while the user dispenses water using the plastic squeeze bottle onto the digital scale.}
    \label{fig:human_exp_setup}
\end{figure}

The experimental setup to collect human demonstrations of bottle squeezing is depicted in Fig. \ref{fig:human_exp_setup}. For the purpose of this paper, data from a single user (one of the authors) was collected. The user was equipped with FSR caps on their thumb and middle finger and sEMGs were placed on the FDS and distal radius bone to measure the FDS muscle activity. The user was instructed to squeeze the bottle with only their thumb and middle finger to dispense water into the cup on the digital scale.  

This experiment compared the dispensed volume of water using the squeeze bottle with a tuned force controller and a tuned adaptive hybrid force-vision controller presented in this work. The tuned values used were $K_{p,\mathrm{f}} = 0.0002$,  $K_{p,\mathrm{f},\mathrm{deform}} = 0.001$, $K_{p,\mathrm{deform}} = 0.003$, $K_d = 0.1$, and $B_{\mathrm{nom}} = 20\%$. Based on the transition zone of 0-0.03mV in the EMG to FSR mapping seen in Fig. \ref{fig:EMGvsFSR}, the hybrid modulation factor was chosen to have the relation $\sigma = max( \frac{0.03 - EMG}{0.03},0)$. During the force controller trials, $\sigma$ was set to 1 to remove any visual components and rely only on grip force feedback. 

 To test the dispensing accuracy and human demonstration replicability of the controller on various bottles, the experiment in this study tested 12 variations of squeeze bottle dispensing cases where the amount of liquid dispensed, bottle size, and number of squeezes per trial were varied. The variations combined dispensing approximately 40mL or 80mL of water, using small (250 mL) or large (500 mL) squeeze bottles, and using 1, 2, or 3 squeezes to dispense the liquid at a rate of approximately 1 mL/s. Two human demonstration of each case were collected and recorded using ROS2 bags. The recorded EMG signals for each of the 24 human demonstrations were then replayed and used as inputs for the force reference of the  adaptive hybrid controller and the force controller. Five trials of each case for each controller were collected.  

\subsection{Results}

\subsubsection{EMG to FSR Mapping}
\begin{figure}
    \centering
    \includegraphics[width=\linewidth]{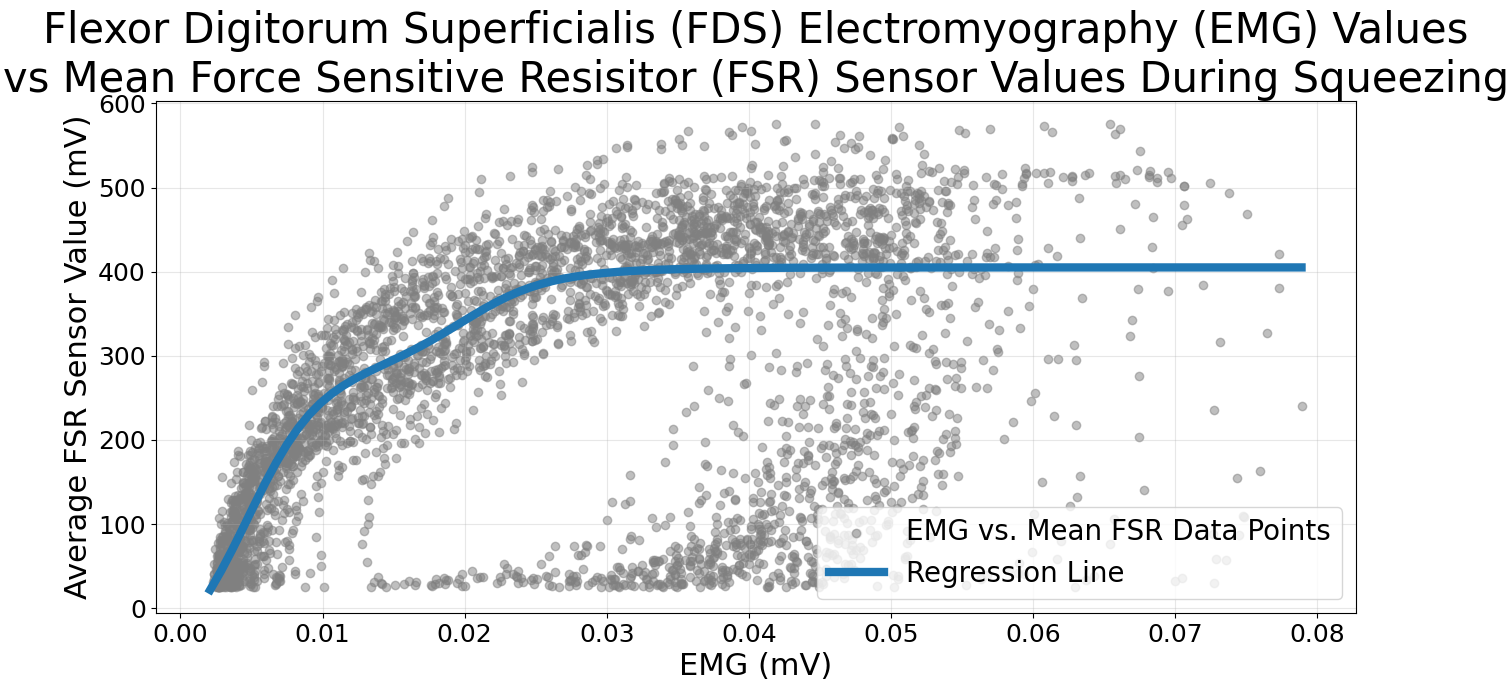}
    \caption{Trained regression of the mean force sensitive resistor (FSR) values as a function of flexor digitorum superficialis (FDS) electromyography (EMG) values during squeezing.}
    \label{fig:EMGvsFSR}
\end{figure}

The data collected to create the mapping between the mean FSR values as a function of FDS EMG values was recorded at 100 Hz with 1 operator performing this squeezing task for 50 trials. For each trial, the operator squeezed the block from the minimum contraction required to hold the block to their maximum voluntary contraction over approximately 3 seconds with 1 second breaks in between. The regression relationship was obtained by training a multilayer perceptron network consisting of two hidden layers with four units each and a single-unit output layer. Sigmoid activation functions were applied to the hidden layers, while the network weights were constrained to be positive using a Softplus transformation. The model was trained using L1 loss, with early stopping applied when the validation loss failed to improve by more than $10^{-6}$ for 2000 consecutive epochs. The collected data points and the resulting regression can be seen in Fig. \ref{fig:EMGvsFSR}. The resultant regression has an L1-loss of 18.9442. This regression is used for the mapping from EMG to desired FSR force values in the controllers of this experiment.

\begin{figure}
    \centering
    \includegraphics[width=0.9\linewidth]{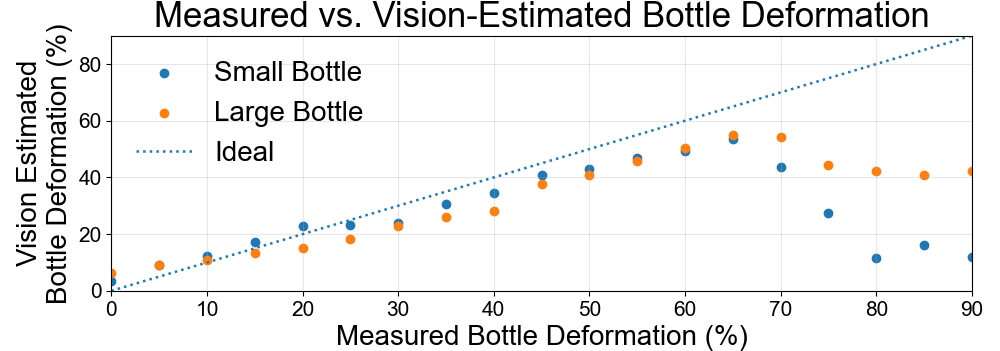}
    \caption{Vision estimated compared to measured bottle deformation from 0-90\% deformation.}
    \label{fig:deformation_est_result}
\end{figure}

\begin{figure*}[t]
    \centering
    \includegraphics[width=\linewidth]{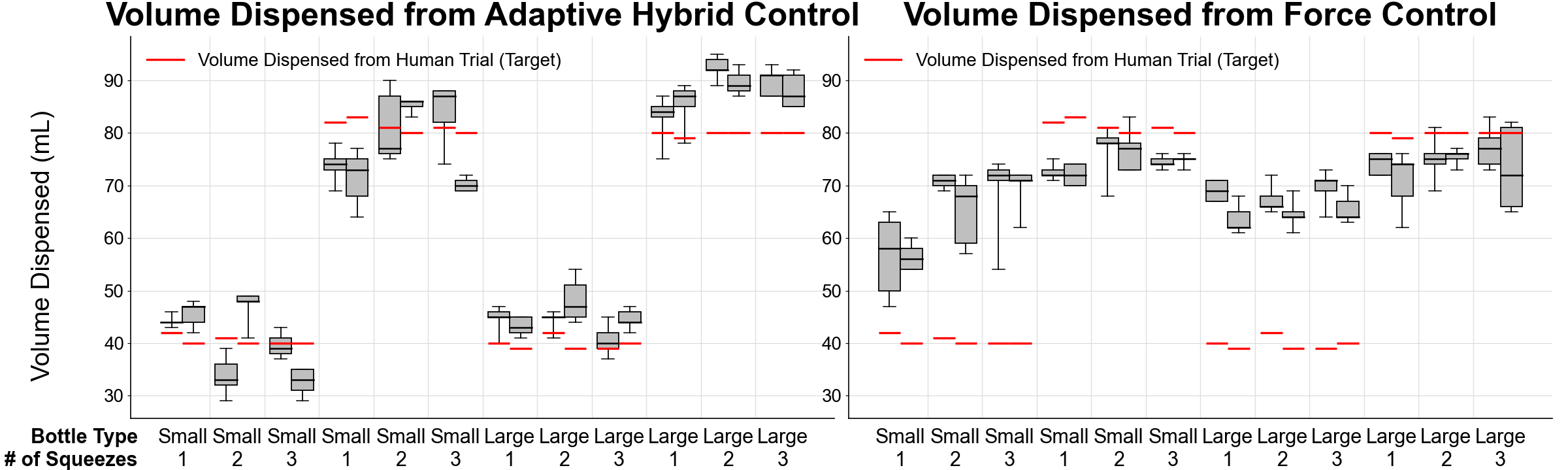}

    \caption{Volume dispensed using adaptive hybrid and force control under different bottle and squeezing conditions. Each candlestick pair represents the five repeated robot trials corresponding to human demonstrations 1 (left) and 2 (right), with red lines indicating the human dispensed volumes.}
    \label{fig:volume_dispensed_comparison}
\end{figure*}

\begin{table*}[t]
\caption{Per-condition dispensing performance of the adaptive hybrid vision-force and force controllers. SD denotes the standard deviation of the five robot dispensing trials per-condition. MAE and RMSE are computed relative to the corresponding human-demonstration dispensed volume. All values are in mL.}
\label{tab:controller_condition_statistics}
\centering
\footnotesize
\setlength{\tabcolsep}{4pt}
\renewcommand{\arraystretch}{1.10}

\begin{tabular*}{\textwidth}{@{\extracolsep{\fill}} c c c c c c c c c c}

\toprule
\multirow{2}{*}{\textbf{Bottle}} &
\multirow{2}{*}{\textbf{Target}} &
\multirow{2}{*}{\textbf{Squeezes}} &
\multirow{2}{*}{\textbf{Human Demonstration}} &
\multicolumn{3}{c}{\textbf{Adaptive Hybrid Control}} &
\multicolumn{3}{c}{\textbf{Force Control}} \\
\cmidrule(lr){5-7}
\cmidrule(lr){8-10}
& & & & \textbf{SD} & \textbf{MAE} & \textbf{RMSE} & \textbf{SD} & \textbf{MAE} & \textbf{RMSE} \\
\midrule

\multirow{12}{*}{Small}
& \multirow{6}{*}{40}
    & \multirow{2}{*}{1} & 1 & 1.10 & 2.20 & 2.41 & 7.89 & 14.60 & 16.22 \\
&   &                     & 2 & 2.51 & 5.60 & 6.03 & 2.61 & 16.40 & 16.57 \\
\cmidrule(lr){3-10}
&   & \multirow{2}{*}{2} & 1 & 3.83 & 7.20 & 7.97 & 1.30 & 29.80 & 29.82 \\
&   &                     & 2 & 3.39 & 7.00 & 7.63 & 6.76 & 25.20 & 25.92 \\
\cmidrule(lr){3-10}
&   & \multirow{2}{*}{3} & 1 & 2.41 & 2.00 & 2.19 & 8.35 & 28.80 & 29.75 \\
&   &                     & 2 & 2.61 & 7.40 & 7.76 & 4.28 & 29.60 & 29.85 \\
\cmidrule(lr){2-10}

& \multirow{6}{*}{80}
    & \multirow{2}{*}{1} & 1 & 3.27 & 8.20 & 8.71 & 1.52 & 9.40 & 9.50 \\
&   &                     & 2 & 5.32 & 11.60 & 12.54 & 2.00 & 11.00 & 11.14 \\
\cmidrule(lr){3-10}
&   & \multirow{2}{*}{2} & 1 & 6.96 & 6.00 & 6.23 & 5.07 & 4.20 & 6.18 \\
&   &                     & 2 & 1.30 & 5.20 & 5.33 & 4.15 & 4.40 & 4.90 \\
\cmidrule(lr){3-10}
&   & \multirow{2}{*}{3} & 1 & 6.02 & 5.60 & 6.07 & 1.14 & 6.60 & 6.68 \\
&   &                     & 2 & 1.30 & 9.80 & 9.87 & 1.10 & 5.20 & 5.29 \\
\midrule

\multirow{12}{*}{Large}
& \multirow{6}{*}{40}
    & \multirow{2}{*}{1} & 1 & 2.70 & 4.60 & 5.20 & 2.00 & 29.00 & 29.06 \\
&   &                     & 2 & 1.79 & 4.20 & 4.49 & 2.88 & 24.60 & 24.73 \\
\cmidrule(lr){3-10}
&   & \multirow{2}{*}{2} & 1 & 1.95 & 2.80 & 2.97 & 2.79 & 25.40 & 25.52 \\
&   &                     & 2 & 4.21 & 9.20 & 9.94 & 2.88 & 25.60 & 25.73 \\
\cmidrule(lr){3-10}
&   & \multirow{2}{*}{3} & 1 & 3.05 & 2.40 & 3.16 & 3.44 & 30.60 & 30.75 \\
&   &                     & 2 & 1.95 & 4.60 & 4.92 & 2.88 & 25.60 & 25.73 \\
\cmidrule(lr){2-10}

& \multirow{6}{*}{80}
    & \multirow{2}{*}{1} & 1 & 4.60 & 4.80 & 4.98 & 2.05 & 5.80 & 6.08 \\
&   &                     & 2 & 4.39 & 6.80 & 7.51 & 5.76 & 8.20 & 9.69 \\
\cmidrule(lr){3-10}
&   & \multirow{2}{*}{2} & 1 & 2.30 & 12.40 & 12.57 & 4.30 & 5.40 & 6.31 \\
&   &                     & 2 & 2.41 & 9.60 & 9.84 & 1.52 & 4.60 & 4.80 \\
\cmidrule(lr){3-10}
&   & \multirow{2}{*}{3} & 1 & 2.68 & 9.80 & 10.09 & 4.02 & 4.00 & 4.56 \\
&   &                     & 2 & 3.32 & 8.00 & 8.53 & 8.04 & 8.00 & 9.90 \\
\bottomrule
\end{tabular*}
\end{table*}

\begin{figure*}[t]
    \centering
    \includegraphics[width=\textwidth]{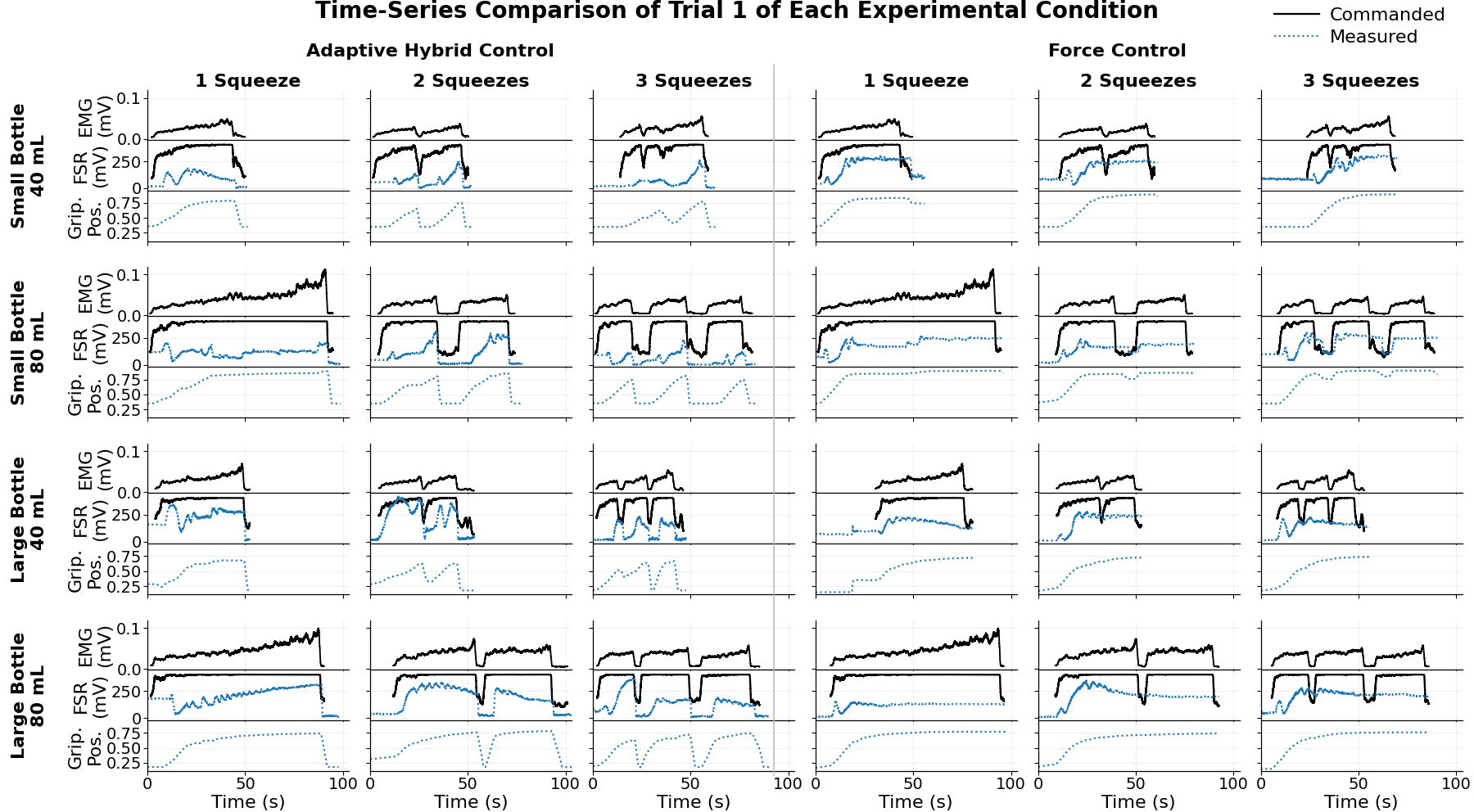}
    \caption{Time-series comparison between the adaptive force-vision and force control methods for trial 1 across all bottle, target volume, and squeezing conditions. The solid black line represents the commanded FSR/EMG value, while the blue dotted lines represent measured signals.}
    \label{fig:trial1_timeseries_comparison}
\end{figure*}

\subsubsection{Bottle Deformation Estimation Using Computer Vision}
The performance of the bottle deformation estimation vision system is evaluated in Fig.~\ref{fig:deformation_est_result}. The vision system accurately estimates bottle deformation within the range of 0--65\% deformation, achieving mean absolute errors (MAEs) of 5.26\% and 6.96\% for the small and large bottles, respectively. However, the estimation accuracy decreases significantly beyond 70\% deformation. At high deformation levels, the visible shape of the bottle changes substantially, causing the deformation estimation to use less deformed areas of the bottle as it cannot detect the fully deformed area between the gripper. As a result, inaccurate deformation estimates are produced in this highly deformed region. Therefore, to ensure reliable deformation feedback during control, the maximum gripper closing position was limited to 0.8, corresponding to approximately 70\% bottle deformation, where the vision estimator maintains sufficient accuracy.

\subsubsection{Adaptive Hybrid Control vs. Force Control}
Fig. \ref{fig:volume_dispensed_comparison} presents the dispensed-volume distributions for the adaptive hybrid and force controllers across all bottle, target-volume, squeeze-count, and human demonstration trial conditions. For each condition, the candlestick plots summarize the five repeated robot trials, while the red horizontal lines indicate the corresponding volume dispensed during the human demonstration. Table \ref{tab:controller_condition_statistics} complements this figure by summarizing the dispensing performance for each condition and human demonstration. The table reports the standard deviation of the five robot dispensing trials, together with the MAE and root mean squared error (RMSE) relative to the corresponding human-demonstration dispensed volume.

Fig. \ref{fig:trial1_timeseries_comparison} shows the time-series responses for both controllers for the first trial of each experimental condition. The figure includes the EMG command, the commanded FSR signal from the EMG to FSR mapping, the measured FSR signal during the bottle squeezing task, and the resultant gripper position from each controller, over time. Only one trial has been presented here to provide a clear visual, but these results represents the general response across all trials.

\section{Discussion}
The adaptive hybrid controller generally maintained the dispensed volume closer to the human-demonstration target, particularly with the 40~mL conditions. Averaged across the 24 condition--human demonstration combinations, the adaptive hybrid controller achieved an MAE of 6.54~mL and an RMSE of 6.96~mL, compared with 15.92~mL and 16.45~mL, respectively, for force control. The difference was most evident for the 40~mL target, for which the mean MAE across conditions was 4.93~mL for adaptive hybrid control and 25.43~mL for force control. Table~\ref{tab:controller_condition_statistics} further shows that the variability across the five repeated robot trials was comparable between the two approaches, with mean per-condition standard deviations of 3.14~mL and 3.70~mL for adaptive hybrid and force control, respectively. Overall, these results indicate that the adaptive hybrid controller reproduced the human-demonstration dispensed volume with greater accuracy while maintaining comparable repeatability.

Further differences between the two controllers can be observed in the time-series responses shown in Fig.~\ref{fig:trial1_timeseries_comparison}. Bottle squeezing presents a unique deformable-object manipulation task in which the relationship between applied force and dispensed volume depends on the residual deformation of the bottle. Specifically, the same squeezing force can produce different dispensed volumes depending on the bottle's deformation state. Partially releasing the bottle between successive squeezes allows air to re-enter and the bottle to recover towards its undeformed state, altering the force and deformation required for subsequent squeezes. Consequently, reproducing the grip relaxation between squeezes is an important component of the human demonstration.

In the two- and three-squeeze conditions, the adaptive hybrid controller reproduced this relaxation behaviour. Decreases in the EMG command between squeezes were accompanied by reopening of the gripper, allowing the bottle to partially recover before the subsequent squeeze. In contrast, this behaviour was not reproduced by the force controller. The force-controlled gripper generally remained closed after the initial squeeze rather than reopening in response to the reduction in the human-derived EMG command. This suggests that force tracking alone was insufficient to reproduce the temporal grip-relaxation behaviour required for repeated manipulation of the deformable bottle.

A similar trend can be observed across the gripper-position responses of the force controller. Across all tested conditions, the gripper typically closed to approximately 0.75 and remained near this position for the remainder of the trial. This provides a possible explanation for the clustering of force-controlled dispensed volumes in the approximately 60--80~mL range observed in Fig.~\ref{fig:volume_dispensed_comparison}. Because the force controller produced similar gripper trajectories across different conditions, the resulting bottle deformation and dispensed volume also tended to be similar despite differences in the demonstrated squeezing behaviour. These results indicate that reproducing the force response alone does not provide sufficient control to capture the condition-dependent manipulation behaviour of the deformable bottle.

Overall, the two controllers produced substantially different squeezing behaviours despite being driven by the same human-derived EMG command profile. The adaptive hybrid controller reproduced the grip-relaxation pattern associated with repeated human squeezing while also achieving lower dispensing errors overall. This demonstrates the benefit of incorporating object deformation information alongside force feedback when reproducing human manipulation of deformable objects.

\section{Conclusion and Future Work}

This work presented an adaptive hybrid force-vision controller that integrates EMG-derived human force references, force feedback from FSRs, and vision-based bottle deformation estimation for deformable bottle manipulation. The proposed controller was experimentally validated across 12 squeezing conditions with varying bottle sizes, target dispensed volumes, and squeezing patterns. Compared with a pure force controller, the proposed approach achieved significantly closer accuracy with human-demonstrated dispensed volumes by adapting the gripper behaviour based on both interaction forces and object deformation. These results demonstrate the importance of incorporating deformation information alongside force feedback for deformable object manipulation. Furthermore, the successful reproduction of EMG-driven human squeezing behaviours highlights the potential of EMGs as a non-invasive modality for collecting natural human demonstrations in contact-rich manipulation tasks, enabling future research in learning from demonstration and behaviour cloning.

Future work will focus on improving the generalizability of the proposed approach for broader laboratory automation scenarios. Variations in bottle materials, liquid viscosity, and object geometry may affect the relationship between applied force and resulting deformation, requiring additional sensing and adaptive mechanisms. Furthermore, extending the approach to other laboratory equipment, including fragile glass containers, will require control strategies that can balance secure grasping with safe contact forces to prevent damage while maintaining reliable manipulation. Ultimately, this research aims to enable robots to acquire human-like manipulation skills for complex laboratory environments through intuitive and non-invasive human demonstrations.

\bibliographystyle{IEEEtran}
\bibliography{Bibliography}

@article{xue2023,
title = {Dynamic gripping force estimation and reconstruction in EMG-based human-machine interaction},
journal = {Biomedical Signal Processing and Control},
volume = {80},
pages = {104216},
year = {2023},
issn = {1746-8094},
doi = {https://doi.org/10.1016/j.bspc.2022.104216},
author = {Jiaqi Xue and King Wai Chiu Lai},
}

@incollection{okafor2022anatomy,
  title={Anatomy, shoulder and upper limb, hand flexor digitorum superficialis muscle},
  author={Okafor, Lauren and Varacallo, Matthew A},
  booktitle={StatPearls [Internet]},
  year={2022},
  publisher={StatPearls Publishing}
}

@ARTICLE{11021574,
  author={Bazina, Tomislav and others},
  journal={IEEE Transactions on Neural Systems and Rehabilitation Engineering}, 
  title={Koopman-Driven Grip Force Prediction Through EMG Sensing}, 
  year={2025},
  volume={33},
  number={},
  pages={2192-2202},
  doi={10.1109/TNSRE.2025.3576110}}

@ARTICLE{9305235,
  author={Cheon, Sangheui and others},
  journal={IEEE Transactions on Medical Robotics and Bionics}, 
  title={Single EMG Sensor-Driven Robotic Glove Control for Reliable Augmentation of Power Grasping}, 
  year={2021},
  volume={3},
  number={1},
  pages={179-189},
  doi={10.1109/TMRB.2020.3046847}}

@article{KS2025104563,
title = {EMG controlled mobile robot equipped with gripper mechanism for fine motor skills training in rehabilitation},
journal = {Results in Engineering},
volume = {26},
pages = {104563},
year = {2025},
issn = {2590-1230},
doi = {https://doi.org/10.1016/j.rineng.2025.104563},
author = {Geetha, {K S} and others}
}

@article{dosen2015emg,
  author    = {Dosen, Strahinja and others},
  title     = {EMG Biofeedback for Online Predictive Control of Grasping Force in a Myoelectric Prosthesis},
  journal   = {Journal of NeuroEngineering and Rehabilitation},
  volume    = {12},
  number    = {1},
  pages     = {55},
  year      = {2015},
  doi       = {10.1186/s12984-015-0047-z},
}

@INPROCEEDINGS{9198439,
  author={Gowtham, S. and others},
  booktitle={International Conference on Wireless Communications Signal Processing and Networking (WiSPNET)}, 
  title={EMG-Based Control of a 5 DOF Robotic Manipulator}, 
  year={2020},
  volume={},
  number={},
  pages={52-57},
  doi={10.1109/WiSPNET48689.2020.9198439}}

@inproceedings{cong2024tacfr,
  title={TacFR-Gripper: A Reconfigurable Fin-Ray-Based Gripper with Tactile Skin for In-Hand Manipulation},
  author={Cong, Qingzheng and Fan, Wen and Zhang, Dandan},
  booktitle={Actuators},
  volume={13},
  number={12},
  pages={521},
  year={2024},
  organization={MDPI}
}

@article{ABCofEMG,
author = {Konrad, Peter},
year = {2005},
month = {01},
pages = {},
title = {The abc of emg},
volume = {1},
journal = {A practical introduction to kinesiological electromyography}
}

@InProceedings{Rusu_ICRA2011_PCL,
	author    = {Radu Bogdan Rusu and Steve Cousins},
	title     = {{3D is here: Point Cloud Library (PCL)}},
	booktitle = {IEEE International Conference on Robotics and Automation},
	year      = {2011},
	address   = {Shanghai, China}
}

@ARTICLE{Zhu_deform,
  author={Zhu, Jihong and others},
  journal={IEEE Robotics \& Automation Magazine}, 
  title={Challenges and Outlook in Robotic Manipulation of Deformable Objects}, 
  year={2022},
  volume={29},
  number={3},
  pages={67-77},
  doi={10.1109/MRA.2022.3147415}}

@INPROCEEDINGS{Wada_ICRA,
  author={Wada, T. and others},
  booktitle={Proceedings. IEEE International Conference on Robotics and Automation}, 
  title={Robust manipulation of deformable objects by a simple PID feedback}, 
  year={2001},
  volume={1},
  number={},
  pages={85-90},
  doi={10.1109/ROBOT.2001.932534}}

@Inbook{Wada2000,
author={Wada, T.},
editor={Henrich, Dominik and W{\"o}rn, Heinz},
title={Indirect Simultaneous Positioning of Extensible Deformable Objects},
bookTitle={Robot Manipulation of Deformable Objects},
year={2000},
publisher={Springer London},
address={London},
pages={73-89},
isbn="978-1-4471-0749-1",
doi={10.1007/978-1-4471-0749-1_6}}

@INPROCEEDINGS{lee_icra,
  author={Lee, Alex X. and others},
  booktitle={ IEEE International Conference on Robotics and Automation (ICRA)}, 
  title={Learning force-based manipulation of deformable objects from multiple demonstrations}, 
  year={2015},
  volume={},
  number={},
  pages={177-184},
  doi={10.1109/ICRA.2015.7138997}}

@ARTICLE{10.3389/frobt.2020.00073,
    
AUTHOR={Sanchez, Jose  and others },
           
TITLE={Blind Manipulation of Deformable Objects Based on Force Sensing and Finite Element Modeling},
          
JOURNAL={Frontiers in Robotics and AI},
          
VOLUME={7},
  
YEAR={2020},
  
DOI={10.3389/frobt.2020.00073},
  
ISSN={2296-9144}}

@INPROCEEDINGS{zhong_robio,
  author={Zhong, Hanzhong and others},
  booktitle={IEEE International Conference on Robotics and Biomimetics (ROBIO)}, 
  title={Regressor-Based Model Adaptation for Shaping Deformable Linear Objects with Force Control}, 
  year={2023},
  volume={},
  number={},
  pages={1-7},
  doi={10.1109/ROBIO58561.2023.10354796}}

@INPROCEEDINGS{maitin-shepard,
  author={Maitin-Shepard, Jeremy and others},
  booktitle={IEEE International Conference on Robotics and Automation}, 
  title={Cloth grasp point detection based on multiple-view geometric cues with application to robotic towel folding}, 
  year={2010},
  volume={},
  number={},
  pages={2308-2315},
  doi={10.1109/ROBOT.2010.5509439}}

@INPROCEEDINGS{sun_icra,
  author={Sun, Li and others},
  booktitle={IEEE International Conference on Robotics and Automation}, 
  title={Accurate garment surface analysis using an active stereo robot head with application to dual-arm flattening}, 
  year={2015},
  volume={},
  number={},
  pages={185-192},
  doi={10.1109/ICRA.2015.7138998}}

@ARTICLE{Arriola-Rios2020, 
AUTHOR={Arriola-Rios, Veronica E.  and others},
TITLE={Modeling of Deformable Objects for Robotic Manipulation: A Tutorial and Review},   
JOURNAL={Frontiers in Robotics and AI},          
VOLUME={7},  
YEAR={2020},     
DOI={10.3389/frobt.2020.00082},  
ISSN={2296-9144},
}

@Article{Gentile2023,
AUTHOR = {Gentile, Cosimo and others},
TITLE = {Manipulation Tasks in Hazardous Environments Using a Teleoperated Robot: A Case Study at CERN},
JOURNAL = {Sensors},
VOLUME = {23},
YEAR = {2023},
NUMBER = {4},
ARTICLE-NUMBER = {1979},
PubMedID = {36850591},
ISSN = {1424-8220},
DOI = {10.3390/s23041979}
}

@article{Cooper2025,
doi={10.1126/scirobotics.adv7932},
author = {Andrew I. Cooper  and others },
title = {Accelerating discovery in natural science laboratories with AI and robotics: Perspectives and challenges},
journal = {Science Robotics},
volume = {10},
number = {106},
pages = {eadv7932},
year = {2025},
doi = {10.1126/scirobotics.adv7932},
eprint = {https://www.science.org/doi/pdf/10.1126/scirobotics.adv7932}}

@INPROCEEDINGS{leite_icra,
  author={Antonio C. Leite and others},
  booktitle={Proceedings IEEE International Conference on Robotics and Automation.}, 
  title={Hybrid vision-force robot control for tasks on unknown smooth surfaces}, 
  year={2006},
  volume={},
  number={},
  pages={2244-2249},
  doi={10.1109/ROBOT.2006.1642037}}

@article{antonio_sage,
author = {Antonio C. Leite and others},
title ={Hybrid Adaptive Vision—Force Control for Robot Manipulators Interacting with Unknown Surfaces},

journal = {The International Journal of Robotics Research},
volume = {28},
number = {7},
pages = {911-926},
year = {2009},
doi = {10.1177/0278364909101932},

eprint = {https://doi.org/10.1177/0278364909101932}
}

@INPROCEEDINGS{sheng_m2vip,
  author={Sheng, Qi and others},
  booktitle={27th International Conference on Mechatronics and Machine Vision in Practice (M2VIP)}, 
  title={Hybrid Vision-Force Robot Force Control for Tasks on Soft Tissues}, 
  year={2021},
  volume={},
  number={},
  pages={705-710},
  doi={10.1109/M2VIP49856.2021.9665090}}

@article{gohel2020review,
  author = {Gohel, Vidhi and Mehendale, Ninad},
  title = {Review on electromyography signal acquisition and processing},
  journal = {Biophysical reviews},
  year = {2020},
  volume = {12},
  number = {6},
  pages = {1361--1367},
  doi = {10.1007/s12551-020-00770-w},
  issn = {1867-2450},
}

@article{coggshall1970emgforce,
  author = {Coggshall, J. C. and Bekey, G. A.},
  title = {{EMG}-force dynamics in human skeletal muscle},
  journal = {Medical and biological engineering},
  year = {1970},
  month = {may},
  volume = {8},
  number = {3},
  pages = {265--270},
  doi = {10.1007/BF02477243},
  issn = {1741-0444},
}

@INPROCEEDINGS{VAIRO,
  author={Lee, Jeffrey and others},
  booktitle={34th IEEE International Conference on Robot and Human Interactive Communication (RO-MAN)}, 
  title={{VAIRO}: A Vision-Based Adaptive Impedance-Control Robotic Framework}, 
  year={2025},
  volume={},
  number={},
  pages={462-468},
  doi={10.1109/RO-MAN63969.2025.11217620}}

@INPROCEEDINGS{VAISI,
  author={Lee, Jeffrey and others},
  booktitle={IEEE-RAS 24th International Conference on Humanoid Robots (Humanoids)}, 
  title={{VAISI}: Vision-Based Adaptive Impedance-Control for Surgical Incisions}, 
  year={2025},
  volume={},
  number={},
  pages={1-8},
  doi={10.1109/Humanoids65713.2025.11203178}}

@article{THUROW2026100373,
title = {Editorial: Robotics in laboratory automation},
journal = {SLAS Technology},
volume = {36},
pages = {100373},
year = {2026},
issn = {2472-6303},
doi = {https://doi.org/10.1016/j.slast.2025.100373},
author = {Kerstin Thurow and others}
}

\end{document}